\documentclass[runningheads]{llncs}

\usepackage[T1]{fontenc}
\usepackage{graphicx}
\usepackage{booktabs}
\usepackage{amsmath}
\usepackage{url}
\usepackage{xcolor}
\usepackage{listings}
\usepackage{tabularx}
\usepackage{array}
\usepackage{ragged2e}
\usepackage[section]{placeins}
\usepackage[hidelinks]{hyperref}

\newcolumntype{Y}{>{\RaggedRight\arraybackslash}X}
\newcolumntype{C}{>{\centering\arraybackslash}X}
\newcolumntype{P}[1]{>{\RaggedRight\arraybackslash}p{#1}}
\newcommand{\code}[1]{\texttt{#1}}

\begin{document}

\title{Revisiting Risky Tackle Detection with Vision Transformers}
\titlerunning{Reproducing ViT-Based Risk Detection in Football}

% -----------------------------------------------------------------------------
% Author block
% -----------------------------------------------------------------------------
%\author{Anonymous Author(s)}
%\authorrunning{Anonymous}
%\institute{Anonymous Institution(s)}

 \author{
 Syed Ahsan Masud Zaidi\inst{1} \and
 Lior Shamir\inst{1} \and
 Scott Dietrich\inst{2}
 }

% % Short author list for the running header.
% % LNCS commonly uses "First Author et al." when there are 3+ authors.
% \authorrunning{S. A. M. Zaidi et al.}

 \institute{
 Kansas State University, Manhattan, KS, USA\\
 \email{\{ahsanzaidi, lshamir\}@ksu.edu}
 \and
 Albright College, Reading, PA, USA\\
 \email{sdietrich@albright.edu}
 }

\maketitle

\begin{abstract}
This paper is a Track 2 reproducibility companion to an ICPR 2026 study
on risky tackle detection in American football practice
videos~\cite{Zaidi_ICPR2026}. The original work fine-tuned a Video Vision
Transformer (ViViT) on 733 clips labeled with the SATT-3 rubric. It used
focal loss, Taguchi $L_{18}$ augmentation, and 5-fold cross-validation.
It reported risky-class recall of 0.67 and risky-class F1 of 0.59. This
companion documents the released artifact and traces those numbers to
specific scripts, fold outputs, and aggregation files. The reproduced
headline is \code{run\_15}. It combines Gaussian noise with static
brightness decrease and uses no rotation and no flip. Its fold-mean
risky recall is 0.667 and its fold-mean risky F1 is 0.588. These values
match the published headline after rounding. The ablation shows that
brightness is the dominant factor. Its risky-recall main-effect range is
0.055, which is larger than the ranges for rotation, flip, and noise.
Without augmentation, ViViT reaches risky recall of 0.545 and does not
exceed the C3D baseline of 0.583. The raw clips show identifiable
student athletes, so they cannot be redistributed. The artifact provides
a public sample for pipeline checks and a controlled route for full-data
review.
\end{abstract}

\section{Introduction}
\label{sec:intro}

Head and neck injuries in American football are strongly linked to poor
tackling form. Practice is where form is taught and corrected. Coaches
cannot review every repetition on film. Risky repetitions can therefore
go unnoticed. Automated detection can rank clips for coach review and
help standardize feedback across sessions. Nafi et al. established the
task with a 3D convolutional network on practice clips~\cite{Nafi2022RiskyTackle3D}.
Follow-up work isolated the relevant athletes through instance
segmentation~\cite{instseg}. The broader goal is a safety pipeline that
works under tight data and compute constraints.

This companion uses a design-of-experiments view. Taguchi orthogonal
arrays provide a compact way to study augmentation factors under a fixed
compute budget~\cite{taguchi}. The tackle task needs this kind of
controlled ablation because the full factorial augmentation space is
large and video training is expensive.

Four constraints make the task hard. The labels are imbalanced, with
64.7\% safe and 35.3\% risky clips. The discriminative signal often
appears near first contact. Training is compute intensive. The footage
shows identifiable student athletes and is governed by an institutional
review protocol.

The original study~\cite{Zaidi_ICPR2026} uses 733 single-athlete tackle clips
recorded against a padded dummy. The set contains 474 safe clips and 259
risky clips. Each clip is trimmed to 32 frames, with 15 frames before
and 16 frames after the first point of contact (FPOC). FPOC localization
is manual in~\cite{Zaidi_ICPR2026}. A zero-shot alternative, GRAZE, was later
developed on the same dataset~\cite{Zaidi_2026_CVPR}. Labels follow the SATT-3 rubric. Scores of 0 or 1 map to risky. Scores of 2 or 3 map to safe.
The model is ViViT~\cite{arnab2021vivit}, initialized from
\code{google/vivit-b-16x2-kinetics400} pretrained on
Kinetics-400~\cite{kinetics}. Inputs are 32 frames at $224\times224$.
Training uses focal loss~\cite{focal} ~\cite{iot}, balanced sampling, 5-fold
stratified cross-validation, and a Taguchi $L_{18}$ augmentation
schedule. The headline is risky recall 0.67 and risky F1 0.59. The C3D
baseline reaches risky recall 0.583 and risky F1 0.560~\cite{Nafi2022RiskyTackle3D}.
The model in this artifact is specifically ViViT. It is not a per-frame
image ViT.

This companion makes five contributions. First, it provides a documented
pipeline that can be tested on a public sample. Authorized users can run
it on the full dataset after local path configuration. Second, it maps
the main reported numbers to scripts, configurations, and output files.
Third, it explains the data-access restriction for the IRB-constrained
clips. Fourth, it states that the heatmap values are the fold-mean
\code{opt\_*} metrics selected with per-fold macro-F1 threshold tuning.
Fifth, it reports a Taguchi $L_{18}$ main-effects analysis and shows
that brightness decrease is the strongest augmentation factor.

The paper is organized as follows. Section~\ref{sec:setup} describes the
artifact and environment. Section~\ref{sec:data} describes data access.
Section~\ref{sec:repro} traces the headline result. Section~\ref{sec:analysis}
reports the reproducibility analysis. Section~\ref{sec:lessons} gives
lessons learned. Section~\ref{sec:conclusion} concludes.

\section{Artifact Overview and Environment}
\label{sec:setup}

\subsection{Repository and Script Inventory}
\label{sec:setup:repo}

The anonymous artifact is available at
\url{https://anonymous.4open.science/r/tacklestudy_vivit-7919/readme.md}.
The repository contains the released code, public-sample utilities, and
environment files. Table~\ref{tab:mapping} maps each result to its code
lineage and output artifact.

\begin{table}[t]
\centering
\caption{Paper-to-code mapping for the released artifact.}
\label{tab:mapping}
\scriptsize
\setlength{\tabcolsep}{2.5pt}
\renewcommand{\arraystretch}{1.12}
\begin{tabularx}{\textwidth}{@{}P{0.28\textwidth}P{0.27\textwidth}P{0.13\textwidth}Y@{}}
\toprule
Paper item & Script or source & Config. & Output artifact \\
\midrule
Headline and Table~\ref{tab:repro} &
\code{vivit\_train\_taguchi.py} plus \code{consolidate\_metrics.py} &
50 epochs &
\code{metrics\_summary.csv}; consolidated heatmap \\
$L_{18}$ main effects in Table~\ref{tab:maineffects} &
heatmap values from \code{opt\_*} metrics &
5 folds &
performance heatmap \\
Complete heatmap audit in Table~\ref{tab:complete_heatmap} &
\code{consolidate\_metrics.py} &
fold means &
\code{consolidated\_metrics.csv} \\
Public-sample build &
\code{Taguchi\_datasets.py} &
demo split &
\code{taguchi\_runs/run\_*/} \\
Full-data fold directories &
prepared \code{taguchi\_runs} &
5-fold $L_{18}$ &
\code{run\_*/fold\_*/} \\
Full-sweep job submission &
SLURM launcher &
20 runs, 5 folds &
job logs \\
Statistical aggregation &
\code{consolidate\_taguchi.py} &
optional &
summary CSV files and figures \\
\bottomrule
\end{tabularx}
\end{table}

The headline lineage uses \code{vivit\_train\_taguchi.py}. The script
sets 32 frames, input size 224, batch size 2, 50 maximum epochs,
\code{google/vivit-b-16x2-kinetics400}, and gradient accumulation over 8
steps. It uses a \code{WeightedRandomSampler}, early stopping, and
macro-F1 threshold tuning. The released script exposes focal-loss
hyperparameters through command-line arguments. Its parser defaults are
$\alpha=0.6$ and $\gamma=1.6$, but the archived SLURM logs for the
reported heatmap record $\alpha=0.55$ and $\gamma=1.3$. We therefore
treat $\alpha=0.55$ and $\gamma=1.3$ as the reproduction settings for
Table~\ref{tab:repro} and Table~\ref{tab:complete_heatmap}. The same
script writes both standard argmax metrics (\code{std\_*}) and
threshold-tuned metrics (\code{opt\_*}) into \code{metrics\_summary.csv}.
The reported heatmap uses the \code{opt\_*} columns.

\code{consolidate\_metrics.py} crawls the result tree, reads each
\code{metrics\_summary.csv}, and writes a consolidated metric table. Its
plotting routine uses the \code{opt\_risky\_recall}, \code{opt\_accuracy},
and \code{opt\_macro\_f1} columns. This confirms that the heatmap lineage
is threshold-tuned, not an argmax-only lineage. \code{Taguchi\_datasets.py}
constructs the public-sample run directories. It implements static
brightness increase and static brightness decrease. The full 733-clip
results are tied to the archived full-data fold directories and archived
metric outputs.

The released scripts keep local path variables for run directories.
Reproducers must update those paths for their own storage. This keeps the
artifact close to the code used in the study while allowing review on a
different machine.

\subsection{Software Environment}
\label{sec:setup:env}

The environment is pinned in \code{Environment.yml}. It uses Python 3.10
with the CUDA 12.1 build of PyTorch 2.2.2. It also includes torchvision
0.17.2, transformers 4.51.3, timm 0.4.12, decord 0.6.0, opencv-python
4.10, numpy 1.23.5, and scikit-learn 1.4.2. Reporting uses pandas,
seaborn, and matplotlib. Setup uses Listing~\ref{lst:env}.

\begin{lstlisting}[caption={Environment setup.},label={lst:env}]
conda env create -f Environment.yml
conda activate vivit_pyt
\end{lstlisting}

Four dependency risks matter. The ViViT checkpoint should be cached if
the training system has limited internet access. The pinned transformers
version was validated with timm 0.4.12. Upgrading timm may change the
model loading path. Decord 0.6.0 may also require a source build on some
Linux systems. Writing \code{taguchi\_summary.xlsx} needs the
\code{openpyxl} engine for pandas. It is not currently pinned in
\code{Environment.yml}, so a fresh install may need it added manually.

\subsection{Hardware and Compute Budget}
\label{sec:setup:hw}

The reported full-data runs used one NVIDIA H100-80GB GPU per job on PSC
Bridges-2 through an ACCESS allocation. The full sweep contains 20 runs
and 5 folds per run, for 100 model trainings. The job launcher processes
three runs per job and loops over folds 0 through 4. The full sweep was
run in batches. The cost was about 72 GPU-hours, or about 3 to 4 hours
per run. Gradient accumulation over 8 steps gives an effective batch size
of 16.

Storage is separate from compute. Each run keeps its own augmented train
and validation clips for every fold. The full Taguchi build occupies
about 700 GB. A reproducer with limited storage should process runs in
batches and archive metrics before deleting completed run directories.

The public-sample workflow is lighter. A single modern CUDA GPU can test
preprocessing, training, evaluation, and aggregation. Lower-memory GPUs
need a smaller batch size. Reducing the frame count below 32 is expected
to reduce risky recall, but that effect was not measured here.

\section{Data Access and IRB Considerations}
\label{sec:data}

The 733 clips show identifiable student athletes during practice. The
institutional review protocol does not permit public redistribution of
the raw clips. This restriction also covers the 178-clip subset first
used in~\cite{Nafi2022RiskyTackle3D}.

The class distribution affects every metric. There are 474 safe clips
(64.7\%) and 259 risky clips (35.3\%). This is a 1.83 to 1 ratio. A
classifier that predicts safe for every clip reaches 64.7\% accuracy but
misses every risky tackle. This failure mode motivates a recall-first
evaluation.

A public sample is available at
\url{https://www.kaggle.com/datasets/ahsanzaidi786/tacklenet-sample}. It
supports environment setup, run-directory construction, preprocessing,
training, evaluation, and metric aggregation. It does not support
quantitative reproduction of the reported cross-validation metrics. Full
reproduction of Table~\ref{tab:complete_heatmap} requires the full
733-clip dataset.

Reviewers who need restricted clips for badge evaluation can request
access during review. Requests are routed through the RRPR chairs and
then to the authors. Each requester signs a data use agreement. The
agreement restricts use to the evaluation of this submission. It also
prohibits redistribution and re-identification. It requires deletion of
the clips after review closes. After the agreement is countersigned, the
data are delivered through a university-managed secure transfer service.
Access is limited to the named requester.

\section{Reproducing the Headline Results}
\label{sec:repro}

Twenty configurations were evaluated. Eighteen use the Taguchi $L_{18}$
augmentation schedule. \code{run\_0} uses oversampling without
transformation. \code{run\_0\_original} uses neither oversampling nor
augmentation. Each configuration is evaluated with 5 stratified folds.
The headline result corresponds to \code{run\_15}. That run combines
Gaussian noise with static brightness decrease and uses no rotation and
no flip. It reaches risky recall 0.667 and risky F1 0.588.

Listing~\ref{lst:repro} gives the verification workflow. The full-data
step requires the authorized 733-clip fold directories.

\begin{lstlisting}[caption={Verification workflow for the released code.},label={lst:repro}]
# Step 1: create the environment
conda env create -f Environment.yml
conda activate vivit_pyt

# Step 2a: verify the public-sample pipeline
python Taguchi_datasets.py \
    --video_root /path/to/public_sample/videos \
    --label_csv  /path/to/public_sample/labels.csv \
    --output_root taguchi_runs

# Step 2b: run one authorized full-data fold
python vivit_train_taguchi.py \
    --runs run_15 \
    --fold 0 \
    --alpha 0.55 \
    --gamma 1.3 \
    --threshold_strategy macro_f1

# Step 2c: run the complete full-data sweep on the cluster
# For the archived heatmap, use the log-recorded focal-loss values.
bash run_vivit_train_taguchi.sh h100-80:1 GPU-shared asc180003p

# Step 3: aggregate the metric files
python consolidate_metrics.py
python consolidate_taguchi.py --results_dir ./taguchi_runs_GRADCAM_RESULTS
\end{lstlisting}

The script reports two metric families. The \code{std\_*} values use the
standard 0.5 threshold through argmax. The \code{opt\_*} values use a
per-fold threshold selected to maximize macro-F1 on that fold. The
published heatmap and the reproduced headline use the \code{opt\_*}
values. This distinction is necessary for exact reproduction.

\begin{table}[t]
\centering
\caption{Training configuration for the reproduced headline lineage.}
\label{tab:config}
\scriptsize
\setlength{\tabcolsep}{4pt}
\renewcommand{\arraystretch}{1.12}
\begin{tabularx}{\textwidth}{@{}P{0.30\textwidth}Y@{}}
\toprule
Setting & Value \\
\midrule
Training script & \code{vivit\_train\_taguchi.py} \\
Backbone & ViViT~\cite{arnab2021vivit}; \code{google/vivit-b-16x2-kinetics400}~\cite{kinetics} \\
Frames and resolution & 32 frames at $224\times224$ \\
Temporal window & 15 frames before FPOC and 16 frames after FPOC \\
Batch and precision & batch size 2; bf16 on supported GPUs \\
Gradient accumulation & 8 steps; effective batch size 16 \\
Learning rate and schedule & $5\times10^{-5}$; cosine schedule with 10\% warmup \\
Weight decay & 0.01 \\
Epochs & 50 maximum epochs with early stopping; patience 10 \\
Loss & focal loss with log-recorded $\alpha_t=0.55$ for risky clips, $0.45$ for safe clips, and $\gamma=1.3$ \\
Sampling & \code{WeightedRandomSampler} \\
Primary threshold & per-fold macro-F1 tuning; stored as \code{opt\_*} metrics \\
Secondary threshold & standard argmax; stored as \code{std\_*} metrics \\
Split source & prepared 5-fold directories with stratified class balance \\
GPU & H100-80GB on PSC Bridges-2 \\
\bottomrule
\end{tabularx}
\end{table}

The current script parser defaults are $\alpha=0.6$ and $\gamma=1.6$.
Those defaults are useful for rerunning the public artifact. They are not
used as the source of the published heatmap in this companion. The
heatmap audit uses the SLURM-log configuration, $\alpha=0.55$ and
$\gamma=1.3$.

\begin{table}[t]
\centering
\caption{Baseline comparison and headline result. Values are 5-fold means from the threshold-tuned heatmap.}
\label{tab:repro}
\scriptsize
\setlength{\tabcolsep}{3pt}
\renewcommand{\arraystretch}{1.08}
\begin{tabularx}{\textwidth}{@{}Yccccc@{}}
\toprule
Configuration & Risky recall & Risky F1 & Risky prec. & Accuracy & Safe recall \\
\midrule
C3D baseline~\cite{Nafi2022RiskyTackle3D} & 0.583 & 0.560 & 0.538 & 0.711 & 0.769 \\
ViViT, no augmentation & 0.545 & 0.530 & 0.522 & 0.660 & 0.724 \\
ViViT, oversample only & 0.537 & 0.540 & 0.547 & 0.679 & 0.757 \\
\midrule
Best Taguchi (\code{run\_15}) & \textbf{0.667} & \textbf{0.588} & 0.535 & 0.669 & 0.670 \\
\midrule
Fig.~4 heatmap in~\cite{Zaidi_ICPR2026} & 0.667 & 0.588 & 0.535 & 0.669 & 0.670 \\
\midrule
Mean $\pm$ std, runs 01 to 18 & $0.572 \pm 0.041$ & $0.550 \pm 0.020$ & $0.541 \pm 0.027$ & $0.672 \pm 0.016$ & $0.726 \pm 0.037$ \\
\bottomrule
\end{tabularx}
\end{table}
\FloatBarrier

The main-paper abstract reports risky recall and risky F1. Fig.~4
in~\cite{Zaidi_ICPR2026} reports all five heatmap metrics for \code{run\_15}. The
row above therefore repeats the full heatmap values for that configuration.

ViViT without augmentation reaches risky recall 0.545 in the archived
heatmap. This rounds to 0.55, not 0.58. The original ICPR narrative text
states this baseline as 0.58, which is close to the C3D value of 0.583.
This companion uses the heatmap value, 0.545, since the heatmap is the
direct source for these tables. Readers checking the original text
should expect this small difference. The best Taguchi configuration
exceeds C3D by 0.084 in risky recall and by 0.028 in risky F1. This
comparison is descriptive. The C3D baseline and the ViViT experiment
were not run under the same controlled protocol.

\section{Reproducibility Analysis}
\label{sec:analysis}

\subsection{Lineage and Aggregation}
\label{sec:analysis:lineage}

The reported heatmap values are the threshold-tuned \code{opt\_*}
values. Each fold writes both standard and optimal metrics to
\code{metrics\_summary.csv}. \code{consolidate\_metrics.py} gathers those
files and creates consolidated summaries. This is why Table~\ref{tab:repro}
and Table~\ref{tab:complete_heatmap} use threshold-tuned fold means.

The practical path requirement is simple. \code{BASE\_RUNS\_DIR} must
point to a directory with the expected \code{run\_*/fold\_*} structure.
\code{RESULTS\_BASE\_DIR} must point to the output tree that stores the
per-fold \code{metrics\_summary.csv} files. Reviewers can check the
public-sample pipeline without the restricted clips.

\subsection{Taguchi $L_{18}$ Factors}
\label{sec:analysis:taguchi}

The ablation varies four augmentation factors. Noise has 2 levels.
Brightness, rotation, and flip have 3 levels each. This gives a full
factorial space of 54 combinations. The Taguchi $L_{18}$ array covers
the same factor set with 18 runs. This reduces the cost from about 160
to 220 GPU-hours to about 54 to 72 GPU-hours. The design estimates main
effects under an orthogonality assumption. It does not estimate factor
interactions.

\begin{table}[t]
\centering
\caption{Taguchi $L_{18}$ factors for the reported 733-clip heatmap. Brightness uses static increase and static decrease levels.}
\label{tab:factors}
\footnotesize
\setlength{\tabcolsep}{4pt}
\renewcommand{\arraystretch}{1.12}
\begin{tabularx}{\textwidth}{@{}P{0.20\textwidth}P{0.10\textwidth}Y P{0.12\textwidth}@{}}
\toprule
Factor & Code & Levels & Count \\
\midrule
Noise & $A$ & None; Gaussian noise & 2 \\
Brightness & $B$ & Static increase; static decrease; same & 3 \\
Rotate & $C$ & Left; right; none & 3 \\
Flip & $D$ & Horizontal; vertical; none & 3 \\
\midrule
Full factorial space & & & 54 \\
$L_{18}$ runs & & & 18 \\
\bottomrule
\end{tabularx}
\end{table}

\begin{table}[t]
\centering
\caption{Taguchi main effects for risky-class recall. Values are computed from the 18 threshold-tuned heatmap recalls.}
\label{tab:maineffects}
\scriptsize
\setlength{\tabcolsep}{3pt}
\renewcommand{\arraystretch}{1.08}
\begin{tabularx}{\textwidth}{@{}P{0.18\textwidth}YYYYP{0.14\textwidth}@{}}
\toprule
Factor & Level 1 & Level 2 & Level 3 & Range & Best \\
\midrule
Noise ($A$) & None: 0.564 & Noise: 0.580 & -- & 0.016 & Noise \\
\textbf{Brightness ($B$)} & Incr.: 0.552 & \textbf{Decr.: 0.607} & Same: 0.559 & \textbf{0.055} & Decrease \\
Rotate ($C$) & Left: 0.558 & Right: 0.574 & None: 0.586 & 0.028 & None \\
Flip ($D$) & Horiz.: 0.582 & Vert.: 0.571 & None: 0.565 & 0.017 & Horizontal \\
\bottomrule
\end{tabularx}
\end{table}

Brightness has the largest range. Decreased brightness is the best
level. A plausible explanation is that reduced luminance suppresses
background clutter and makes the contact region more salient. Rotation
is second. No rotation is the best rotation level. Large geometric
changes can disturb orientation cues that matter for tackle form. Noise
and flip have smaller effects. The best observed run combines noise with
brightness decrease and uses no rotation and no flip.

\begin{table}[!t]
\centering
\caption{Complete heatmap values used for the reproduction check. Values are fold means.}
\label{tab:complete_heatmap}
\scriptsize
\setlength{\tabcolsep}{4pt}
\renewcommand{\arraystretch}{1.05}
\begin{tabular}{lccccc}
\toprule
Configuration & Accuracy & Risky prec. & Risky recall & Risky F1 & Safe recall \\
\midrule
C3D baseline & 0.711 & 0.538 & 0.583 & 0.560 & 0.769 \\
No augmentation & 0.660 & 0.522 & 0.545 & 0.530 & 0.724 \\
Oversampled no augmentation & 0.679 & 0.547 & 0.537 & 0.540 & 0.757 \\
\code{run\_01} & 0.674 & 0.536 & 0.560 & 0.544 & 0.736 \\
\code{run\_02} & 0.673 & 0.552 & 0.498 & 0.521 & 0.768 \\
\code{run\_03} & 0.670 & 0.535 & 0.563 & 0.542 & 0.728 \\
\code{run\_04} & 0.645 & 0.507 & 0.594 & 0.544 & 0.673 \\
\code{run\_05} & 0.655 & 0.510 & 0.560 & 0.526 & 0.707 \\
\code{run\_06} & 0.674 & 0.542 & 0.587 & 0.557 & 0.721 \\
\code{run\_07} & 0.639 & 0.490 & 0.509 & 0.497 & 0.709 \\
\code{run\_08} & 0.651 & 0.509 & 0.626 & 0.560 & 0.665 \\
\code{run\_09} & 0.678 & 0.545 & 0.583 & 0.560 & 0.730 \\
\code{run\_10} & 0.699 & 0.581 & 0.539 & 0.557 & 0.786 \\
\code{run\_11} & 0.667 & 0.535 & 0.575 & 0.550 & 0.718 \\
\code{run\_12} & 0.680 & 0.543 & 0.575 & 0.558 & 0.737 \\
\code{run\_13} & 0.678 & 0.553 & 0.600 & 0.564 & 0.720 \\
\code{run\_14} & 0.672 & 0.539 & 0.633 & 0.577 & 0.694 \\
\code{run\_15} & \textbf{0.669} & 0.535 & \textbf{0.667} & \textbf{0.588} & 0.670 \\
\code{run\_16} & 0.695 & 0.585 & 0.543 & 0.556 & 0.779 \\
\code{run\_17} & 0.674 & 0.544 & 0.551 & 0.544 & 0.741 \\
\code{run\_18} & 0.701 & 0.599 & 0.541 & 0.559 & 0.789 \\
\bottomrule
\end{tabular}
\end{table}

\FloatBarrier
\subsection{Stability of Comparisons}
\label{sec:analysis:stability}

Across the 18 $L_{18}$ runs, risky recall spans 0.498 to 0.667. The
standard deviation is 0.041. Risky F1 spans 0.497 to 0.588. Its standard
deviation is 0.020. F1 is more stable than recall. Augmentation mainly
changes the recall and precision tradeoff. Those shifts partly cancel in
F1.

The comparison against C3D should be read with care. \code{run\_15}
exceeds the C3D baseline by 0.084 recall and 0.028 F1. The comparison is
not fully controlled. The two systems were trained on different dataset
sizes and under different protocols. This companion documents the
conditions under which the reported result was obtained. It does not
claim a definitive superiority margin.

Folds are assigned at the clip level using stratified splitting. The
same athlete can appear in both training and validation folds. This can
make performance estimates optimistic. Athlete-stratified or
session-stratified splits are the recommended next step.

\subsection{Experimental Protocol and Generalization}
\label{sec:analysis:protocol}

Each clip is trimmed to a 32-frame window. The window contains 15 frames
before the annotated FPOC and 16 frames after it. Raw clips run 200 to
1500 frames at 30 fps before trimming. Frames are resized to
$224\times224$. BGR frames are converted to RGB before the ViViT
processor. FPOC localization is described in~\cite{Zaidi_ICPR2026} and automated
in~\cite{Zaidi_2026_CVPR}.

The label schema follows the SATT-3 rubric. Scores 0 and 1 map to risky,
label 1. Scores 2 and 3 map to safe, label 0. Domain experts annotated
all 733 clips. The final distribution is 474 safe clips and 259 risky
clips.

Augmentation is applied only to training splits after fold construction.
Validation folds contain original clips only. The risky class is targeted
so that training approaches class balance. A subset of safe clips is also
augmented. This reduces the chance that the model separates classes by
augmentation artifacts alone. The quantitative results are tied to the
archived full-data fold directories and the archived metric outputs.

The generalization scope is narrow. The dataset comes from one
institution with fixed recording conditions. Performance may degrade
under different camera angles, standoff distances, helmet styles, pad
styles, field surfaces, and athlete body types. Clip-level folds may also
inflate recall relative to athlete-stratified or session-stratified
evaluation.

\section{Lessons Learned}
\label{sec:lessons}

The readme, license, and environment file were reconstructed during
artifact preparation. Rebuilding an environment from a live codebase is
slow and error prone. Recording dependency versions at project start is
much cheaper.

Metric lineage must be explicit. The heatmap values are threshold-tuned
\code{opt\_*} metrics. A reproducer who uses standard argmax metrics will
not obtain the same table. The artifact therefore writes both metric
families and names them in the output CSV files.

A shared configuration file would make future audits easier. The main
settings should be captured in one machine-readable file. Scripts should
load that file and apply explicit command-line overrides. This would
reduce ambiguity about focal-loss weights, sampler behavior, and
threshold selection. The focal-loss discrepancy between script defaults
and SLURM logs is the clearest example.

Reviewer data access needs early planning. A data use agreement and a
secure transfer route take time to set up. Teams working with
identifiable participant footage should plan this route during data
collection, not after acceptance.

The $L_{18}$ design covered a 54-combination factor space with only 18
model trainings. It saved about 108 to 144 GPU-hours compared with an
exhaustive search. The main-effects analysis still isolated a clear
factor: brightness decrease.

\section{Conclusion}
\label{sec:conclusion}

The headline result of~\cite{Zaidi_ICPR2026} is reproduced from the released
artifact lineage. \code{run\_15} reaches risky recall 0.667 and risky F1
0.588. These values match the reported 0.67 and 0.59 after rounding. The
artifact documents the training script, the threshold-tuned metric files,
and the aggregation path behind the heatmap. Brightness decrease is the
strongest augmentation factor. ViViT without augmentation does not
exceed the C3D baseline in risky recall. Clip-level fold construction is
the main protocol limitation. Athlete-stratified or session-stratified
evaluation is the recommended next step.

\bibliographystyle{splncs04}
\bibliography{References}

\end{document}